\pdfoutput=1

\documentclass[11pt]{article}

\usepackage[]{EACL2023}

\usepackage{times}
\usepackage{latexsym}

\usepackage[T1]{fontenc}

\usepackage[utf8]{inputenc}

\usepackage{microtype}

\usepackage{graphicx}
\usepackage{multirow}
\usepackage{amsmath}

\usepackage{inconsolata}

\newcommand{\ourmodel}{{\textsc{inspect}}}
\newcommand{\ourmodelgen}{{\textsc{inspect-gen}}}
\newcommand{\Sref}[1]{\S\ref{#1}}

\newcommand{\ignore}[1]{}

\title{Unsupervised Keyphrase Extraction via Interpretable Neural Networks} 

\author{
  Rishabh Joshi$^\clubsuit$~\thanks{\hspace{0.5em}Equal Contribution}\hspace{0.5em} \quad Vidhisha Balachandran$^\clubsuit$~\footnotemark[1]\hspace{0.5em} \quad Emily Saldanha$^\diamondsuit$\\ \quad \bf{Maria Glensk}i$^\diamondsuit$
  \quad \bf{Svitlana Volkova}$^\diamondsuit$ \quad
  \bf{Yulia Tsvetkov}$^\spadesuit$\\
  \vspace{0.5mm}$^\clubsuit$Language Technologies Institute, Carnegie Mellon University \\
$^\diamondsuit$Pacific Northwest National Laboratory \\
$^\spadesuit$Paul G.~Allen School of Computer Science \& Engineering, University of Washington \\
        \tt \{rjoshi2,vbalacha\}@cs.cmu.edu,\\ \tt\{emily.saldanha,maria.glenski,svitlana.volkova\}@pnnl.gov,\\ \tt yuliats@cs.washington.edu
}
\begin{document}
\maketitle
\begin{abstract}
Keyphrase extraction aims at automatically extracting a list of ``important'' phrases representing the key concepts in a document. Prior approaches for unsupervised keyphrase extraction resorted to heuristic notions of phrase importance via embedding clustering or graph centrality, requiring extensive domain expertise. Our work presents a simple alternative approach which defines keyphrases as document phrases that are salient for predicting the topic of the document. To this end, we propose \ourmodel{}---an approach that uses self-explaining models for identifying influential keyphrases in a document by measuring the predictive impact of input phrases on the downstream task of the document topic classification. We show that this novel method not only alleviates the need for ad-hoc heuristics but also achieves state-of-the-art results in unsupervised keyphrase extraction in four  datasets across two domains: scientific publications and news articles.\footnote{Code: \url{https://github.com/rishabhjoshi/inspect}.} 
\end{abstract}

\section{Introduction}
Keyphrase extraction is crucial for processing and analysis of long documents in specialized (e.g., scientific, medical) domains \cite{application_of_keyphrase1,application_of_keyphrase3, application_of_keyphrase4}. 
The task is challenging, as the notion of phrase importance is context- and domain-dependent.
Therefore, developing domain-agnostic keyphrase annotation guidelines and curating representative hand-labeled datasets is not feasible. This motivates the need for generalizable unsupervised approaches to keyphrase extraction.

Unsupervised keyphrase extraction methods have used heuristic notions of phrase importance  \cite{textrank, autophrase1, yake}. Popular proxies for phrase importance include phrase clustering based on statistical features like word density \cite{tfidf_scoring, yake} and structural features like graph centrality \cite{topicrank, Ding2022AGRankAG} or more recently neural embedding clustering techniques \cite{embed_rank,zhang-etal-2022-mderank,ding-luo-2021-attentionrank,Sun2020SIFRankAN}. However, such methods do not generalize to new domains as they require experts to carefully construct domain-specific heuristics \cite{clinical_phrase_mining}. 

\begin{figure}[t]
\centering
\includegraphics[width=0.85\columnwidth]{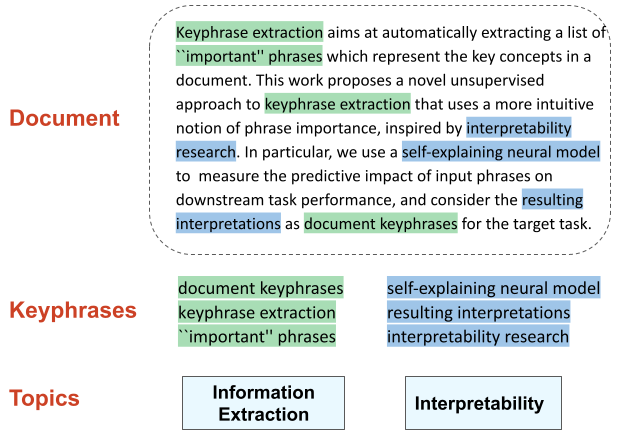}
\caption{
A comprehensive set of keyphrases should highlight important phrases for all major topics in a document. \ourmodel{} identifies such keyphrases using interpretable neural models by measuring how use phrases are for predicting the topic of a text.}
\vspace{-0.2cm}
\label{fig:motivation}
\end{figure}


Historically, topic models \cite{Blei2001LatentDA, Blei2007SupervisedTM, Wallach2006TopicMB} have relied on salient words and phrases in a document, which are similar to the notion of keyphrases, although to the best of our knowledge there is no prior work that identified keyphrases using topic models. In this work, we hypothesize that end-to-end neural models for topic classification  latently rely on salient phrases for document representation and topic classification. Consequently, if we can interpret model decisions via highlighting salient and influential features (phrases) used for neural topic prediction, we can identify such keyphrases. 

Inspired by this intuition, we propose \ourmodel---a novel and simple framework to identify keyphrases by leveraging interpretable text classifiers to highlight phrases important for predicting the topics in a text. Specifically, we adapt an interpretable classifier SelfExplain \cite{self_explain} to jointly predict the topic of an input document and to identify the salient phrases influencing the prediction. The model is distantly supervised using topic labels from off-the-shelf topic-models, eliminating the need for any human/expert annotations. We consider SelfExplain's output interpretations as keyphrases for the input document~(\Sref{sec:framework}). 

\ourmodel{} can be trained on documents of any domain without keyphrase annotations and using distant topic supervision, making them easily adaptable to new domains. 
We contribute two versions of our method: i) \emph{\ourmodel{}}---
individual models trained for topic-classification for each target dataset.
ii) \emph{\ourmodelgen{}}---a more general model pre-trained on a large in-domain  corpus, without finetuning on pre-specified target datasets.

We evaluate \ourmodel{} and \ourmodelgen{} on four benchmark datasets across two domains: scientific documents and news articles (\Sref{sec:exp_setup}). Our results in~\Sref{sec:results} show that \ourmodel{} improves keyphrase extraction performance over strong baselines by 0.8\% F1 on average, without any domain-specific processing. \ourmodelgen{} further improves the performance, outperforming the state of the art in unsupervised keyphrase extraction on 3 out of 4 datasets by 2.7\% F1 on average. Our experiments suggest that \ourmodelgen{} has strong generalization capabilities, and can be used out-of-the-box without finetuning on individual datasets.
Importantly, \ourmodel{} alleviates the need for heuristics and expert-labelled annotations, and thus can be applied to a wide range of domains and problems where keyphrase extraction is important. Our results confirm that the latent keyphrases obtained from an interpretable model correlate with human annotated keyphrases, opening new avenues for research on interpretable models for information extraction.

\section{The \ourmodel{} Framework}
\label{sec:framework}
The goal of the \ourmodel{} framework is to extract important keyphrases in long documents. 
Following the hypothesis that neural text classifiers latently leverage important keyphrases for predicting topics in text, \ourmodel{} extracts keyphrases through interpreting topic classification decisions.
It builds upon an interpretable model, SelfExplain \cite{self_explain}, which learns to attribute text classification decisions to relevant phrases in the input. However, SelfExplain was designed and tested in supervised settings and for single-sentence classification; in this work we explore its extension to unsupervised keyphrase extraction from long documents.  
In what follows, we describe the base SelfExplain model (\Sref{subsec:back}) and the distant supervision setup for \emph{topic classification} (\Sref{subsec:model_topic}).
We outline the training mechanism to jointly predict topics and highlight salient phrases in the document as model interpretations (\Sref{subsec:model}) and finally extract the resulting phrase interpretations as important keyphrases in the document (\Sref{subsec:inf}). 

\subsection{Base Interpretable Model}
\label{subsec:back}
Feature attribution methods for model interpretability include two predominant approaches, (i) post-hoc interpretations of a trained model \cite{hierarchical_attribution,posthocexplain,shapley, Ribeiro2016WhySI}, and (ii) intrinsically (by-design) interpretable models \cite{alvarez2018towards, self_explain}. We adopt the latter approach, specifically SelfExplain~\cite{self_explain} as our phrase attribution model, as the model directly produces interpretations, though in principal any phrase based interpretability techniques could be employed. 

SelfExplain augments a pre-trained transformer-based model (RoBERTa \cite{roberta} in our case) with a local interpretability layer (LIL) and a global interpretability layer (GIL) which are trained to produce local (relevant features from input sample) and global (relevant samples from training data) interpretations respectively. The model can be trained for any text classification tasks using gold task supervision, and produces local and global interpretations along with model predictions. Since our goal is to identify important phrases from the input sample, we use only the LIL layer. The LIL layer takes an input sentence and a set of candidate phrases and quantifies the contribution of a particular phrase for prediction through the activation difference \cite{shrikumar, montavon} between the phrase and sentence representations. 

\subsection{Keyphrase Relevance Model}
\label{subsec:model}
SelfExplain is designed to process single sentences and uses all the phrases spanning non-terminals in a constituency parser as units (candidate phrases) for interpretation. This is computationally expensive for our use-case. To facilitate long document topic classification, we instead define the set of noun phrases (NPs) as the interpretable units, which aligns with prior work in keyphrase extraction of using noun phrases as initial candidate phrases \cite{autophrase1,textrank,topicrank}. \ourmodel{} splits a long document into constituent passages, extracts NPs as candidates, and attributes the contribution of each NP for predicting the topics covered in the passage. 

For each text block $\boldsymbol{X}$ in the input document, we preprocess and identify a set of candidate phrases $\boldsymbol{CP}_X = {\boldsymbol{cp}_1, \boldsymbol{cp}_2, ..., \boldsymbol{cp}_N}$ where $N$ is the number candidate phrases in $\boldsymbol{X}$. From the base RoBERTa model, we obtain contextual [CLS] representations of the entire text block $\mathbf{h}_\textit{[CLS]}$ and individual tokens. We compute phrase representations ${\mathbf{h}_{1}...\mathbf{h}_{N}}$ for each candidate by taking the sum of the RoBERTa representations of each token in the phrase. 

To compute the relevance of each phrase, we construct a representation of the input without the contribution of the phrase, $\mathbf{z}_i$, using the activation differences between the two representations. We then pass it to a classifier layer in the local interpretability module to obtain the label distribution for prediction.
\begin{equation}
    \mathbf{z}_i = g(\mathbf{h}_i) - g(\mathbf{h}_\textit{[CLS]}); ~~~ \boldsymbol{\ell}_i = f(\mathbf{W}^T\mathbf{z}_i + \mathbf{b})
\end{equation}
where $g$ is the ReLU activation function and $W$ and $b$ are the weights and bias of the classifier. Here $\boldsymbol{\ell}_i$ denotes the label distribution obtained on passing the phrase-level representations $\mathbf{z}_i$ through a classification layer $f$ which is either the sigmoid or the softmax function depending on the prediction task (multi-label versus multi-class). We denote the label distribution from the base RoBERTa model for predicting the output using the whole input block as $\boldsymbol{\ell}_\textit{[CLS]}$. We train the model using the cross entropy loss $\mathcal{L}_y$ with respect to the multi-label gold topics $\boldsymbol{Y}_i$ for instance $i$ and an explanation specific loss $\mathcal{L}_e$ using the mean of all phrase-level label distributions such that $\boldsymbol{\ell}_e = \sum_{i=1}^P \boldsymbol{\ell}_i$.
\begin{equation}
    \mathcal{L}_y = -\sum_{j=1}^N \mathbf{y}_j \log (\boldsymbol{\ell}_\textit{[CLS]}), \mathcal{L}_e = -\sum_{j=1}^N \mathbf{y}_j \log (\boldsymbol{\ell}_e)
\end{equation}
The classifier is regularized jointly with $\alpha$ regularization parameter\footnote{$\alpha=0.5$} using explanation and classification loss: $\mathcal{L} = (1-\alpha)\mathcal{L}_y + \alpha \mathcal{L}_e$.

\subsection{Inference}
\label{subsec:inf}
During inference, for each predicted label $y \in \boldsymbol{Y}$, where $\boldsymbol{Y}$ denotes set of all predicted labels for input text $\boldsymbol{X}$, \ourmodel{} calculates an importance score $r_i^y$ with respect to the predicted label $y$ using the difference between the label distribution $\boldsymbol{\ell}_i^y$ for a candidate phrase $\boldsymbol{cp}_i$ and the one obtained using the entire input $\boldsymbol{\ell}_\textit{[CLS]}^y$ as $r_i^y = \boldsymbol{\ell}_\textit{[CLS]}^y - \boldsymbol{\ell}_i^y.$

This score denotes the influence of a candidate keyphrase on the predicted topic. This score denotes the influence of a phrase on the predicted topic---the closer $\boldsymbol{\ell}_i^y$ is to $\boldsymbol{\ell}_\textit{[CLS]}^y$ the less important phrase $i$ is for predicting the topic.
Since the relevance scores are computed with respect to a particular predicted topic and it's label distribution, the scores for the same input are not comparable across different predicted topics in multi-label classification  (since label distributions can vary in magnitude). To aggregate important keyphrases across all predicted topics, we pick the ones that positively impact prediction for each topic (having a positive influence score) as a set of keyphrases.
\[\boldsymbol{KP}(x) = [\boldsymbol{cp}_i\mkern9mu\forall{}\mkern6mur_i^y > 0; y \in \boldsymbol{Y}; i \in \{1:N\}]\]

\begin{table*}[t]
\centering
\small
\begin{tabular}{l|l|l|c|c|c}
    \textbf{Dataset} & \textbf{Type} & \textbf{Split} & \textbf{Total docs} & \textbf{Avg words per doc} & \textbf{Avg keyphrases per doc} \\
    \hline
    \multirow{3}{*}{SciERC} & \multirow{3}{*}{Scientific} & Train  & 350 & 130 & 16\\
    & & Dev & 50 & 130 & 16\\
    & & Test & 100 & 134 & 17\\
    \hline
    \multirow{3}{*}{SciREX} & \multirow{3}{*}{Scientific} & Train  & 306 & 5601 & 353\\
    & & Dev & 66 & 5484 & 354\\
    & & Test & 66 & 6231 & 387\\
    \hline
    \multirow{3}{*}{SemEval17} & \multirow{3}{*}{Scientific} & Train  & 350 & 160 & 21\\
    & & Dev & 50 & 193 & 27\\
    & & Test & 100 & 186 & 23\\
    \hline
    \multirow{3}{*}{500N-KPCrowd} & \multirow{3}{*}{News} & Train  & 400 & 430 & 193\\
    & & Dev & 50 & 465 & 86\\
    & & Test & 50 & 420 & 116\\
    \hline
    BBC News & News & All  & 2225 & 385 & -\\
    ICLR & Scientific & All  & 8317 & 6505 & -\\
\end{tabular}
\caption{Description about the datasets. Average words and keyphrases per document are rounded to the nearest whole number. ICLR and BBC News are used in \ourmodelgen{} setting for training and don't have any labelled keyphrase data.}
\label{tab:datasets}
\end{table*}

\subsection{Distant Supervision via Topic Prediction}
\label{subsec:model_topic}
Obtaining annotations for keyphrases in specialized domains is challenging for supervised keyphrase extraction \cite{clinical_phrase_mining}. Instead, we train the interpretable model in a distant supervision setup for multi-class topic classification and use model interpretations to identify keyphrases, without any keyphrase annotations. Topical information about a document are known to be essential for identifying diverse keyphrases \cite{topicrank, Sterckx2015WhenTM}. Further, a comprehensive set of keyphrases should represent the various major topics in the document to be useful for different long document applications \cite{Liu2010AutomaticKE}. We hypothesize that by using
topic classification as our end-task, our model will learn to highlight---via interpretations it is designed to provide---important and diverse keyphrases in the input document. 

While certain domains like news articles have extensive datasets with human annotated topic labels, others like scientific articles or legal documents require significant effort for human annotation. \ourmodel{} can be trained using annotated topic labels when they exist. In other domains where such annotations are scarce, \ourmodel{} can be trained using labels extracted unsupervisedly using topic models \cite{topic_model}. Experiments in \Sref{sec:results} show results using both settings.

\section{Experimental Setup}
\label{sec:exp_setup}
\subsection{Evaluation Datasets}
\label{subsec:datasets}
We evaluate \ourmodel{} in two domains using four popular keyphrase extraction datasets---scientific publications (SemEval-2017 \cite{semeval2017}, SciERC \cite{scierc}, SciREX \cite{scirex}) and news articles (500N-KPCrowd \cite{news}). 
Dataset details and statistics are shown in Table~\ref{tab:datasets}.

\subsection{Topic Labels}
\label{subsec:topic}
We create distant supervision for \ourmodel{} by labeling the above datasets using document topics as labels. 
We leverage existing topic annotations when such annotations exist. In the 500N-KPCrowd news based dataset, we use existing topic labels (tags or categories such as Sports, Politics, Entertainment) in a one-class classification setting. For the scientific publications domain, we use topic models \cite{topic_model} to extract $T = 75$ topics where each document can be labeled with multiple topics. The scientific domain datasets are trained in a multi-label classification setup.

\subsection{Training Data and Settings}
We train \ourmodel{} in two settings:  
\begin{enumerate}
    \item \textbf{\ourmodel{}} - Here we assume availability of training documents for each of our datasets. We train the model for topic prediction using only the documents and topic labels from the training set of each dataset obtained using the approach outlined in \Sref{subsec:topic}). The training data in this setting, is most closely aligned to the test data, as the documents are of the same topic distribution.
    \item \textbf{\ourmodel{}-Gen} - We assume no access to training documents and train the model on a large external set of documents of a similar domain (ICLR papers for scientific, BBC News for news) but not necessarily of similar topic distribution as the test data (eg. SemEval-2017 has Physics papers). We use ICLR OpenReview dataset with topics obtained using off-the-shelf topic modeling \footnote{\url{https://github.com/gregversteeg/corex_topic}} for the scientific domain and BBC News corpus \cite{bbc_news} with pre-labelled topics for the news domain. 
\end{enumerate}

The model from each setting is then evaluated on the held-out test data of each evaluation dataset.

For the external data, we collect over 8,317 full papers from ICLR and obtained 75 topic labels using topic modeling\footnote{\url{https://github.com/gregversteeg/corex_topic}}. We removed 22 topic labels that were uninformative (list in Appendix Table \ref{tab:removed_topics}) and used the rest to train our model in a multi-label classification setup. The BBC News corpus \cite{bbc_news} consists of 2,225 news article documents, each annotated with one of five topics (business, entertainment, politics, sport, or tech).

We pre-process each document (for training and inference) by splitting it into text blocks of size 512 tokens, where consecutive blocks overlap with a stride size of 128. Following \citet{autophrase1}, for each block we consider all Noun Phrases (NPs) as candidate phrases and extract them using a Noun Phrase extractor from the Berkeley Neural Parser\footnote{\url{https://pypi.org/project/benepar/}}. All hyperparameters were chosen based on development set performance on SciERC. Our final models were trained with a batch size of 8 a learning rate of 2e-5 for 10 epochs.The classification layer dimension was 64 and $\alpha$ was 0.5. We provide more implementation details, including hyperparameter search in Appendix \Sref{app:impl}.

\subsection{Baselines}
\label{sec:baselines}
We compare our method against seven unsupervised keyphrase extraction techniques --- TF-IDF \cite{tfidf_scoring}, TopicRank \cite{topicrank}, Yake \cite{yake}, AutoPhrase \cite{autophrase1,autophrase2}, UKE-CCRank \cite{liang-etal-2021-unsupervised}, MDERank (BERT)\footnote{As of Oct 2022, the authors have not released their model.} \cite{zhang-etal-2022-mderank} and SifRank \cite{Sun2020SIFRankAN}. Out of the chosen baselines, Yake, TF-IDF and AutoPhrase are statistical, TopicRank is graph-based and SifRank, UKE-CCRank and MDERank are neural embedding based methods. For \ourmodel{} setting, we compare with baselines that only use training data documents---TF-IDF, TopicRank, Yake, AutoPhrase, UKE-CCRank and MDERank. For the \ourmodelgen{} setting, we compare with TF-IDF and AutoPhrase trained on our external corpora and SifRank which uses the external corpora to obtain prior likelihood scores for the phrases.

Following prior work and task guidelines \cite{semeval2017, scirex}, \ourmodel{} produces \textbf{span level} keyphrases and distinguishes each occurrence of a keyphrase. In contrast, methods like SifRank, AttentionRank, UKE-CCRank and MDERank are phrase level keyphrase extractors which don't provide span level outputs. To maintain common evaluation, we adapt these methods to span level keyphrase extraction by matching each output keyphrase to all occurrences of the phrase in the document. As our method applies a cutoff on relevance scores and picks any phrase with a positive relevance score as a keyphrase, we cannot be directly compared with baselines which rank candidate phrases and pick top-K phrases as important. To establish a fair setting for evaluation, we choose the average of the number of keyphrase predictions from our model as the 'K' across all baselines.

\begin{table}[t]
\centering
\small
\setlength\tabcolsep{3.5 pt} 
\begin{tabular}{l|l|c|c|c}
 ~ & ~ & \multicolumn{3}{c}{\textbf{F1 Score}}\\
    \textbf{Dataset} & \textbf{Method} & \textbf{Micro} & \textbf{Macro} & \textbf{Weighted}\\
    \hline
    \multirow{2}{*}{SciERC} & RoBERTa & 0.842 & 0.651 & 0.767 \\
    & \ourmodel & 0.836 & 0.658 & \textbf{0.771}\\
    \hline
    \multirow{2}{*}{SciREX} & RoBERTa & 0.609 & 0.404 & 0.641 \\
    & \ourmodel & 0.628 & 0.442 & \textbf{0.697}\\
    \hline
    \multirow{2}{*}{SemEval17} & RoBERTa & 0.819 & 0.613 & 0.731 \\
    & \ourmodel & 0.822 & 0.611 &\textbf{ 0.744}\\
    \hline
    \multirow{2}{*}{500N-KPCrowd} & RoBERTa & 0.916 & 0.880 & 0.910 \\
    & \ourmodel & 0.938 & 0.904 & \textbf{0.939}\\
    \hline
    \hline
    \multirow{2}{*}{ICLR} & RoBERTa & 0.729 & 0.456 & 0.699 \\
    & \ourmodel & 0.743 & 0.492 & \textbf{0.733}\\
    \hline
    \multirow{2}{*}{BBC News} & RoBERTa & 0.880 &	0.851 &	0.876\\
    & \ourmodel & 0.902 &	0.886 &	\textbf{0.894}\\
\end{tabular}
\caption{Proxy Task (Topic prediction) performance. 
Our \ourmodel{} method outperforms a strong RoBERTa baseline on Micro, Macro and Weighted F1 scores. }
\label{tab:topic_prediction_pretraining}
\end{table}
\begin{table*}[t]
\centering
\small
\begin{tabular}{l|l|c|c|c}
\textbf{Dataset} & \textbf{Method} & \textbf{Exact Match F1} & \textbf{Partial Match F1} & \textbf{Avg Exact Partial F1}\\
\hline
\multirow{5}{*}{SciERC} & TF-IDF & 0.0627 & 0.2860 & 0.1743\\
 & TopicRank & 0.2533 & \textbf{0.5680} & 0.4110\\
 & Yake & 0.2230 & 0.5125 & 0.3678\\
 & AutoPhrase & 0.0961 & 0.3145 & 0.2053\\
 & UKE CCRank & \textbf{0.3584} & 0.4804 & 0.4194\\
 & MDERank & 0.3092 & 0.5102 & 0.4097\\
 & \ourmodel & 0.3108 & 0.5524 & \textbf{0.4316}\\
\hline
\multirow{5}{*}{SciREX} & TF-IDF& 0.1521 & 0.3690 & 0.2605\\
 & TopicRank & 0.2298 & 0.4122 & 0.3210\\
 & Yake& 0.1840 & 0.3734 & 0.2787 \\
 & AutoPhrase & 0.1814 & \textbf{0.4236} & 0.3025\\
  & UKE CCRank & 0.0419 & 0.0759 & 0.0589\\
  & MDERank & 0.1241 & 0.3776 & 0.2509\\
 & \ourmodel & \textbf{0.2397} & 0.4127 & \textbf{0.3262}\\
\hline
\multirow{5}{*}{SemEval17} & TF-IDF & 0.0610 & 0.2698 & 0.1654 \\
 & TopicRank& 0.2240 & 0.4312 & 0.3276\\
 & Yake & 0.1687 & 0.3644 & 0.2665\\
 & AutoPhrase & 0.0790 & 0.3404 & 0.2097\\
  & UKE CCRank & 0.2427 & 0.345 & 0.2938\\
  & MDERank & 0.2529 &	0.4818 & 0.3673\\
 & \ourmodel & \textbf{0.2594} & \textbf{0.5185} & \textbf{0.3889}\\
\hline
\multirow{5}{*}{500N-KPCrowd} & TF-IDF & 0.1034 & 0.3520 & 0.2277\\
 & TopicRank & 0.1060 & 0.2346 & 0.1703\\
 & Yake & 0.1380 & 0.3551 & 0.2465\\
 & AutoPhrase & 0.1590 & 0.3608 & 0.2599\\
 & UKE CCRank & \textbf{0.1729} & 0.2873 & 0.2303\\
 & MDERank &0.1522 &	\textbf{0.4197} &	\textbf{0.2859}\\
 & \ourmodel & 0.1608 & 0.3920 & 0.2764\\
\end{tabular}
\caption{Span-match results for unsupervised keyphrase extraction
across datasets in the \ourmodel{} setting. Best performance is indicated in Bold. \textbf{Our model ourperforms baselines on average of exact and partial F1 scores.}
}
\label{tab:span_match_results_in_domain}
\end{table*}

\begin{table*}[t]
\small
\centering
\begin{tabular}{l|l|c|c|c}
\textbf{Dataset} & \textbf{Method} & \textbf{Exact Match F1} & \textbf{Partial Match F1} & \textbf{Avg Exact Partial F1}\\
\hline
\multirow{3}{*}{SciERC} & TF-IDF & 0.2162 & 0.4434 & 0.3298\\
 & AutoPhrase & 0.2416 & 0.6130 & 0.4273\\
 & SifRank & 0.2248 & \textbf{0.735}7 & 0.4803\\
 & \ourmodelgen{} & \textbf{0.4371} & 0.7114 & \textbf{0.5743}\\
\hline
\multirow{3}{*}{SciREX} & TF-IDF& 0.1780 & 0.4008 & 0.2894\\
 & AutoPhrase & 0.2583 & \textbf{0.4993} & \textbf{0.3788}\\
 & SifRank & 0.1234 & 0.3957 & 0.2595\\
 & \ourmodelgen{} & \textbf{0.2601} & 0.4893 & 0.3747\\
\hline
\multirow{3}{*}{SemEval17} & TF-IDF & 0.1810 & 0.3398 & 0.2604 \\
 & AutoPhrase & 0.1104 & 0.4874 & 0.2989\\
 & SifRank & 0.2804 & \textbf{0.6336} & 0.4570\\
 & \ourmodelgen{} & \textbf{0.3246} & 0.6218 & \textbf{0.4732}\\
\hline
\multirow{3}{*}{500N-KPCrowd} & TF-IDF & 0.1398 & 0.3578 & 0.2488\\
 & AutoPhrase & 0.1701 & 0.3918 & 0.2805\\
 & SifRank & \textbf{0.1847} & 0.4125 & \textbf{0.2986}\\
 & \ourmodelgen{} & 0.1776 & \textbf{0.4194} & \textbf{0.2985}\\
\end{tabular}
\caption{Span-match results for unsupervised keyphrase extraction in \ourmodelgen{} (trained on ICLR and BBC News corpus). Best performance is indicated in Bold. \textbf{\ourmodel{} outperforms most baselines}.}
\label{tab:span_match_results_out_domain}
\end{table*}

\subsection{Evaluation Metrics}
\label{sec:evaluation_metrics}
\paragraph{Topic Prediction Evaluation: } To ensure high-quality interpretations from our model, it is imperative that it performs well on topic prediction. We first evaluate \ourmodel{}'s performance on topic prediction using micro, macro, and weighted F1 score of the classifier's predictions compared to true labels across all labels.

\paragraph{Keyphrase Extraction Evaluation:} For our primary evaluation of keyphrase extraction, we evaluate using span match of our predictions and the true labels (human annotated keyphrases). In addition to measuring quality of keyphrases, this evaluation also measures the quality of explanations from our interpretable topic model by measuring how well the keyphrases extracted by \ourmodel{} align with human annotated keyphrases. Prior works \cite{autophrase1,kpminer,topicrank} have mainly focused on \textit{exact match} performance. However, a recent survey highlights that the measure is highly restrictive \cite{survey_keyphrase_extraction} as
simple variations in preprocessing can misalign phrases giving an inaccurate representation of the model's capabilities \cite{Boudin2016HowDP}.

Alternatively, \textit{partial span match} using the word level overlap between the predicted and gold span ranges, has also been explored \cite{partial_match}. But, it is sometimes lenient in scoring.
\citet{survey_keyphrase_extraction} suggest \textit{average of the exact and partial matching} as an appropriate metric based on empirical studies.
Therefore, we evaluate performance using the average of the exact and partial match F1 scores between predicted and true phrases keyphrases.

\section{Results}
\label{sec:results}


\subsection{Topic Prediction with \ourmodel}
First, we compare \ourmodel's effectiveness in classifying the topics with the corresponding non-interpretable encoder baseline, using micro, macro, and weighted F1 score of the classifier's predictions compared to gold standard annotations.
The results in Table \ref{tab:topic_prediction_pretraining} show that our approach outperforms a strong RoBERTa \cite{roberta} baseline for topic prediction across all of our evaluation datasets. 
The difference is more pronounced in larger datasets (SciREX, ICLR, and BBC News), and strong performance on the topic classification task provides confidence that highlighted interpretations are for relevant and major topics in the text.


\subsection{Keyphrase Span Match Performance}
Next, we study the utility of \ourmodel{} in highlighting keyphrases via model interpretations. 
The results for \ourmodel{} are detailed in Table \ref{tab:span_match_results_in_domain} and, for \ourmodelgen{} in Table \ref{tab:span_match_results_out_domain}. 

Results in Table \ref{tab:span_match_results_in_domain} show that even with access to only training set of documents from each dataset, on 3 out of 4 datasets \ourmodel{} outperforms all baselines with $\sim$0.8 average F1 improvements.
In the news domain (500-KPCrowd dataset) \ourmodel{} performs comparably to prior best method. \ourmodel{} has low exact match scores but higher partial match scores indicating misalignments between predicted and gold spans. Additionally, 500N-KPCrowd annotates all instances of a keyphrase as a reference span which favours phrase level methods like AttentionRank in the current evaluation setup. In SciREX, we observe very poor performance of UKE CCRank as it ranks common phrases like ``image'', ``label'', ``method'', etc, very high. 
  

In the \ourmodelgen{} setting, with access to a larger dataset of external documents, our model outperforms prior methods in 3 out of 4 datasets with $\sim$2.7 points average F1 improvements. In the 500N-KPCrowd dataset, \ourmodel{} performs comparably to SifRank with improved Partial Match F1. As Table \ref{tab:span_match_results_out_domain} illustrates, we notice that the model consistently performs better in the \ourmodelgen{} setting when compared with the \ourmodel{} setting, showing that the method benefits from more training data. We particularly see large improvements over the \ourmodel{} setting in the scientific datasets, showing that training on a larger set of documents helps generalize the model in this setting. Our results further show that variations in topic distribution between training and test data don't significantly impact results. \ourmodel{} can thus benefit from large unlabeled documents from similar domains to improve results.

\ourmodel{} improves performance in settings with human annotated topics (news) as well as when topics are extracted using unsupervised topic modeling (scientific). Additionally, most baselines rely on carefully constructed pre- and post-processing to eliminate common phrases and produce high-quality candidates \cite{liang-etal-2021-unsupervised, ding-luo-2021-attentionrank, Sun2020SIFRankAN}. In contrast, \ourmodel{} achieves competitive results without domain expertise and processing for extracting quality keyphrases. Therefore, \ourmodel{} can be easily adapted to new domains without human annotations for topics and with minimal domain knowledge, as we show across two domains.

Our results demonstrate that phrase attribution techniques from interpretability literature can be leveraged to identify high-quality document keyphrases by measuring predictive impact of input phrases on topic prediction. These results also show that our interpretable model in \ourmodel{} produces high quality keyphrases as phrase explanations which correlate with human annotated keyphrases, evaluating the interpretablity aspect of our framework. Crucially, as these keyphrases correlate with human annotated keyphrases, our results validate our initial hypothesis that neural models latently use document keyphrases for tasks like topic classification.  
\begin{table}[t]
\centering
\small
\begin{tabular}{l|c|c}
 ~ & \multicolumn{2}{c}{\textbf{Recall}}\\
    \textbf{Type} & \textbf{Exact} & \textbf{Partial}\\
    \hline
     Metric & 60.65 & 78.34\\ 
     Task & 58.27 & 90.45\\ 
     Material & 72.17 & 86.69\\ 
     Scientific Term & 78.87 & 95.13\\ 
     Method & 65.31 & 95.41\\ 
     Generic & 63.16 & 86.06\\
\end{tabular}
\caption{Exact and partial span match recall scores for different types of keyphrases on the SciERC dataset.}
\label{tab:type_wise}
\end{table}
\section{Discussion}
Here, we present an analysis on the common error types in \ourmodel{} and discuss the strengths and weaknesses of \ourmodel{} using qualitative examples.

\begin{figure*}[t]
\centering
\includegraphics[width=0.85\textwidth]{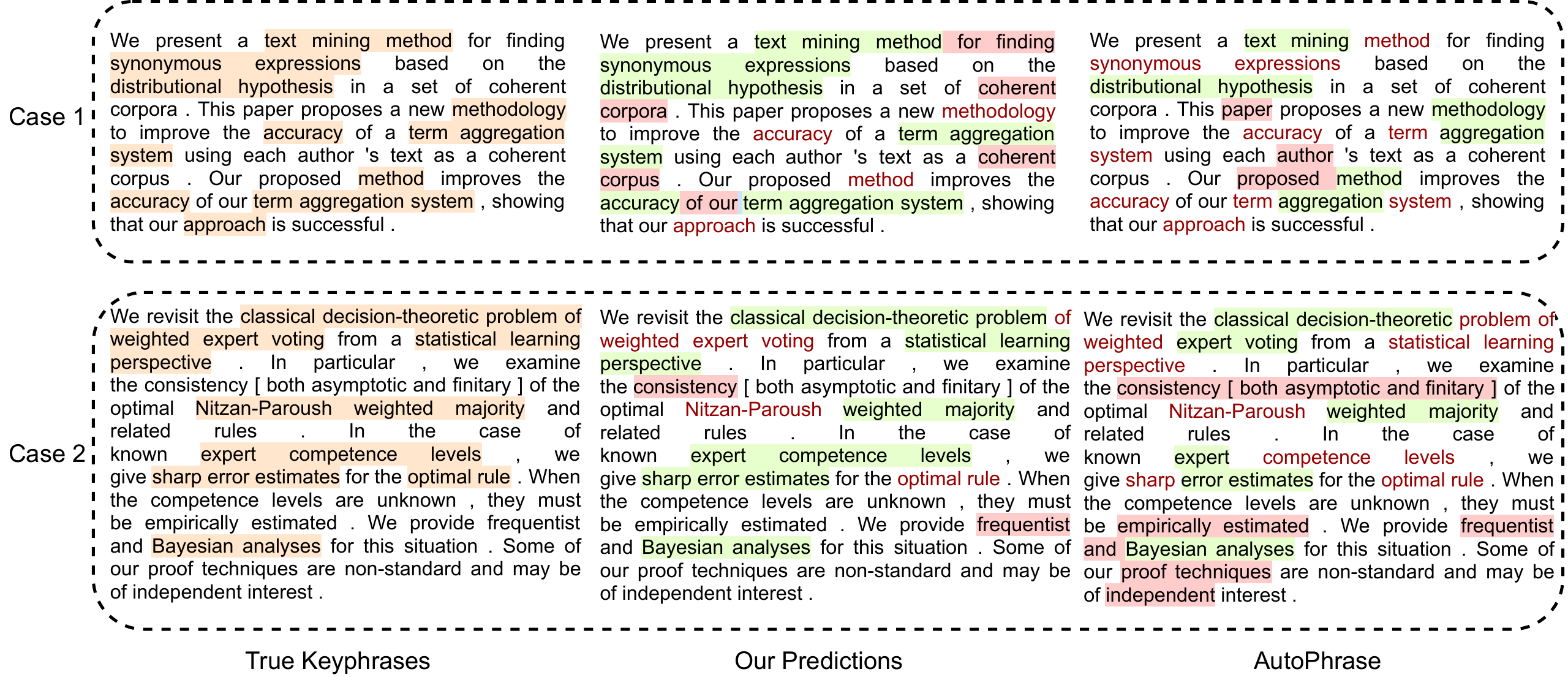} 
\caption{Two data points randomly chosen from the SciERC dataset. Orange spans represent gold standard annotations. Green spans in the predictions represent correctly predicted spans, whereas red spans are spans wrongly predicted as being keyphrases and red text are keyphrases that the model did not identify.}
\label{fig:example}
\end{figure*}

\paragraph{Entity Type Analysis:} We leverage the entity type information in SciERC to observe the performance of \ourmodel{} on specific types of keyphrases. From Table \ref{tab:type_wise}, we see that \ourmodel{} performs best on keyphrases labelled as \textit{Scientific Terms} and \textit{Materials}. 
\textit{Generic} phrases and \textit{Metrics} are usually not representative of topical content, and thus, our method performs poorly on them.
On manual analysis, we noticed that many phrases marked as \textit{Task} are very unique and infrequent, making them harder to identify. 
A high partial match recall but a low exact match recall for \textit{Method} type suggest that many predicted keyphrases are misaligned with the gold labels. 
We believe that alternative downstream tasks can be explored in future to help tailor our approach to capture specific types of entities, based on application requirements. 

\paragraph{Qualitative Analysis} In Figure \ref{fig:example} we show two randomly selected abstracts from the SciERC dataset.
We see that \ourmodel{} tends to extract longer phrases compared to AutoPhrase, which tends to extract mostly unigrams or bigrams.
Overall, our approach is able to extract more relevant phrases than the baseline.
Both \ourmodel{} and AutoPhrase tend to miss generic phrases like `approach' (e.g., as seen in case 1).
Case 2 also demonstrates the \ourmodel{}'s ability TO predict complete phrases, like `classical decision-theoretic problem', instead of AutoPhrase's prediction -- `classical decision-theoretic' which is incomplete. 
From both these examples, we see that \ourmodel is usually able to correctly extract Scientific Terms, and struggles to extract Generic phrases and Metrics. This can be attributed to the usage of topic models to extract the content's topical information.

\section{Related Work} 
Unsupervised keyphrase extraction is typically treated as a ranking problem, given a set of candidate phrases \cite{autophrase1,yake,tfidf_scoring}. 
Broadly, prior approaches can be categorized as statistical, graph-based, embedding-based, or language model based methods;  
\citet{survey_keyphrase_extraction} provide a detailed survey.

Statistical methods exploit notions of information theory directly. Common approaches include TF-IDF based scoring \cite{tfidf_scoring} of phrases with other co-occurrence statistics to enhance performance \cite{key_cluster,kpminer}. 
\citet{yake} shows the importance of incorporating statistical information of the context of each phrase to improve performance.
Statistical approaches typically treat different instances of a phrase equally, which is a limitation.

Graph-based techniques, on the other hand, broadly aim to form a graph of candidate phrases connected based on similarity to each other. Then core components of the graph are chosen as key phrases.
Amongst these, PageRank \cite{page_rank} and TextRank \cite{textrank} assign scores to nodes based on their influence. A common extension is to use weights on the edges denoting the strength of connection \cite{single_rank, rake, topicrank}. 
Position Rank \cite{position_rank} and SGRank \cite{sgrank} combine the ideas from statistical, word co-occurrence and positional information. 
Some approaches, especially applied in the scientific document setting, make use of citation graphs \cite{cite_text_rank, single_rank}, and external knowledge bases \cite{wikirank} to improve keyphrase extraction. In this work, we focus our approach on a general unsupervised keyphrase extraction setting applicable to any domain where such external resources may not be present.

Finally, embedding based techniques \cite{embed_rank, reference_vector_algo,zhang-etal-2022-mderank} make use of word-document similarity using word embeddings \cite{Sun2020SIFRankAN,liang-etal-2021-unsupervised}, while language-model based techniques use word prediction uncertainty to decide informativeness \cite{ngram_lm_keyphrase}. \citet{ding-luo-2021-attentionrank} uses attention scores to calculate phrase importance with the document in an unsupervised manner. 

\section{Conclusion and Future Work}
In this work, we introduced 
\ourmodel{}, a novel approach to unsupervised keyphrase extraction. Our framework  uses a neural model that explains text classification decisions to extract keyphrases via phrase-level feature attribution. Using four standard datasets in two domains, we show that \ourmodel{} outperforms prior methods and establishes state-of-art results in 3 out of 4 datasets.. Through qualitative and quantitative analysis, we show that \ourmodel{} can produce high-quality and relevant keyphrases. \ourmodel{} presents applications of interpretable models beyond explanations for humans. 

\section{Limitations}
Our method uses model explanations for each predicted topic to highlight keyphrases in text. A direct limitation of this method is that our importance scoring is topic-specific and cannot be used to provide an overall rank across topics. Our method therefore cannot provide a ranked list of top-5 or top-10 keyphrases as often done in prior work. While this is a limitation, our current technique of producing a set of all predicted keyphrases is useful in domains like scientific articles where keyphrases are used for downstream applications. Further, as our method produces topic-specific keyphrases, it could potentially miss some keyphrases which are not associated to any predicted topic. Therefore, our approach is beneficial in settings where topic prediction is accurate and feasible to ensure high quality and good coverage of keyphrases.
Finally, this work was also limited by the specific choice of the downstream task - namely, topic prediction. Other downstream tasks, like summarization, can potentially help us gain additional insights from attribution.

\section*{Acknowledgements}
We would like to thank Justin Lovelace, Sachin Kumar, Shangbin Feng, Dheeraj Rajagopal and other members of Tsvetshop Lab for feedback on the paper. This research is supported in part by the Office of the Director of National Intelligence (ODNI), Intelligence Advanced Research Projects Activity (IARPA), via the HIATUS Program contract \#2022-22072200004. 
This material is also funded by the DARPA CMO under Contract No. HR001120C0124 and by the United States Department of Energy (DOE) National Nuclear Security Administration (NNSA) Office of Defense Nuclear Nonproliferation Research and Development (DNN R\&D) Next-Generation AI research portfolio. 
The views and conclusions contained herein are those of the authors and should not be interpreted as necessarily representing the official policies, either expressed or implied, of ODNI, IARPA, or the U.S. Government. The U.S. Government is authorized to reproduce and distribute reprints for governmental purposes notwithstanding any copyright annotation therein.

\bibliography{anthology,custom}

\begin{thebibliography}{52}
\expandafter\ifx\csname natexlab\endcsname\relax\def\natexlab#1{#1}\fi

\bibitem[{Alvarez-Melis and Jaakkola(2018)}]{alvarez2018towards}
David Alvarez-Melis and Tommi~S Jaakkola. 2018.
\newblock Towards robust interpretability with self-explaining neural networks.
\newblock \emph{Neurips}.

\bibitem[{Augenstein et~al.(2017{\natexlab{a}})Augenstein, Das, Riedel,
  Vikraman, and McCallum}]{semeval2017}
Isabelle Augenstein, Mrinal Das, Sebastian Riedel, Lakshmi Vikraman, and Andrew
  McCallum. 2017{\natexlab{a}}.
\newblock \href {https://doi.org/10.18653/v1/S17-2091} {{S}em{E}val 2017 task
  10: {S}cience{IE} - extracting keyphrases and relations from scientific
  publications}.
\newblock In \emph{Proceedings of the 11th International Workshop on Semantic
  Evaluation ({S}em{E}val-2017)}, pages 546--555, Vancouver, Canada.
  Association for Computational Linguistics.

\bibitem[{Augenstein et~al.(2017{\natexlab{b}})Augenstein, Das, Riedel,
  Vikraman, and McCallum}]{augenstein-etal-2017-semeval}
Isabelle Augenstein, Mrinal Das, Sebastian Riedel, Lakshmi Vikraman, and Andrew
  McCallum. 2017{\natexlab{b}}.
\newblock \href {https://doi.org/10.18653/v1/S17-2091} {{S}em{E}val 2017 task
  10: {S}cience{IE} - extracting keyphrases and relations from scientific
  publications}.
\newblock In \emph{Proceedings of the 11th International Workshop on Semantic
  Evaluation ({S}em{E}val-2017)}, pages 546--555, Vancouver, Canada.
  Association for Computational Linguistics.

\bibitem[{Bennani-Smires et~al.(2018)Bennani-Smires, Musat, Hossmann,
  Baeriswyl, and Jaggi}]{embed_rank}
Kamil Bennani-Smires, Claudiu Musat, Andreea Hossmann, Michael Baeriswyl, and
  Martin Jaggi. 2018.
\newblock Simple unsupervised keyphrase extraction using sentence embeddings.
\newblock \emph{arXiv preprint arXiv:1801.04470}.

\bibitem[{Betti et~al.(2020)Betti, Reynaert, Ossenkoppele, Oortwijn, Salway,
  and Bloem}]{application_of_keyphrase3}
Arianna Betti, Martin Reynaert, Thijs Ossenkoppele, Yvette Oortwijn, Andrew
  Salway, and Jelke Bloem. 2020.
\newblock \href {https://doi.org/10.18653/v1/2020.coling-main.586} {Expert
  concept-modeling ground truth construction for word embeddings evaluation in
  concept-focused domains}.
\newblock In \emph{Proceedings of the 28th International Conference on
  Computational Linguistics}, pages 6690--6702, Barcelona, Spain (Online).
  International Committee on Computational Linguistics.

\bibitem[{Blei and McAuliffe(2007)}]{Blei2007SupervisedTM}
David~M. Blei and Jon~D. McAuliffe. 2007.
\newblock Supervised topic models.
\newblock In \emph{NIPS}.

\bibitem[{Blei et~al.(2001)Blei, Ng, and Jordan}]{Blei2001LatentDA}
David~M. Blei, A.~Ng, and Michael~I. Jordan. 2001.
\newblock Latent dirichlet allocation.
\newblock \emph{J. Mach. Learn. Res.}, 3:993--1022.

\bibitem[{Boudin et~al.(2016)Boudin, Mougard, and Cram}]{Boudin2016HowDP}
Florian Boudin, Hugo Mougard, and Damien Cram. 2016.
\newblock How document pre-processing affects keyphrase extraction performance.
\newblock In \emph{NUT@COLING}.

\bibitem[{Bougouin et~al.(2013)Bougouin, Boudin, and Daille}]{topicrank}
Adrien Bougouin, Florian Boudin, and B{\'e}atrice Daille. 2013.
\newblock \href {https://aclanthology.org/I13-1062} {{T}opic{R}ank: Graph-based
  topic ranking for keyphrase extraction}.
\newblock In \emph{Proceedings of the Sixth International Joint Conference on
  Natural Language Processing}, pages 543--551, Nagoya, Japan. Asian Federation
  of Natural Language Processing.

\bibitem[{Brin and Page(1998)}]{page_rank}
Sergey Brin and Lawrence Page. 1998.
\newblock \href {http://www-db.stanford.edu/~backrub/google.html} {The anatomy
  of a large-scale hypertextual web search engine}.
\newblock \emph{Computer Networks}, 30:107--117.

\bibitem[{Campos et~al.(2018)Campos, Mangaravite, Pasquali, Jorge, Nunes, and
  Jatowt}]{yake}
Ricardo Campos, V{\'\i}tor Mangaravite, Arian Pasquali, Al{\'\i}pio~M{\'a}rio
  Jorge, C{\'e}lia Nunes, and Adam Jatowt. 2018.
\newblock A text feature based automatic keyword extraction method for single
  documents.
\newblock In \emph{European conference on information retrieval}, pages
  684--691. Springer.

\bibitem[{Danesh et~al.(2015)Danesh, Sumner, and Martin}]{sgrank}
Soheil Danesh, Tamara Sumner, and James~H. Martin. 2015.
\newblock \href {https://doi.org/10.18653/v1/S15-1013} {{SGR}ank: Combining
  statistical and graphical methods to improve the state of the art in
  unsupervised keyphrase extraction}.
\newblock In \emph{Proceedings of the Fourth Joint Conference on Lexical and
  Computational Semantics}, pages 117--126, Denver, Colorado. Association for
  Computational Linguistics.

\bibitem[{Ding and Luo(2021)}]{ding-luo-2021-attentionrank}
Haoran Ding and Xiao Luo. 2021.
\newblock \href {https://doi.org/10.18653/v1/2021.emnlp-main.146}
  {{A}ttention{R}ank: Unsupervised keyphrase extraction using self and cross
  attentions}.
\newblock In \emph{Proceedings of the 2021 Conference on Empirical Methods in
  Natural Language Processing}, pages 1919--1928, Online and Punta Cana,
  Dominican Republic. Association for Computational Linguistics.

\bibitem[{Ding and Luo(2022)}]{Ding2022AGRankAG}
Haoran Ding and Xiao Luo. 2022.
\newblock Agrank: Augmented graph-based unsupervised keyphrase extraction.
\newblock In \emph{AACL}.

\bibitem[{El-Beltagy and Rafea(2009)}]{kpminer}
Samhaa~R. El-Beltagy and Ahmed Rafea. 2009.
\newblock \href {https://doi.org/https://doi.org/10.1016/j.is.2008.05.002}
  {Kp-miner: A keyphrase extraction system for english and arabic documents}.
\newblock \emph{Information Systems}, 34(1):132--144.

\bibitem[{Florescu and Caragea(2017{\natexlab{a}})}]{tfidf_scoring}
Corina Florescu and Cornelia Caragea. 2017{\natexlab{a}}.
\newblock A new scheme for scoring phrases in unsupervised keyphrase
  extraction.
\newblock In \emph{European Conference on Information Retrieval}, pages
  477--483. Springer.

\bibitem[{Florescu and Caragea(2017{\natexlab{b}})}]{position_rank}
Corina Florescu and Cornelia Caragea. 2017{\natexlab{b}}.
\newblock Positionrank: An unsupervised approach to keyphrase extraction from
  scholarly documents.
\newblock In \emph{Proceedings of the 55th Annual Meeting of the Association
  for Computational Linguistics (Volume 1: Long Papers)}, pages 1105--1115.

\bibitem[{Gallagher et~al.(2017)Gallagher, Reing, Kale, and
  Ver~Steeg}]{topic_model}
Ryan~J Gallagher, Kyle Reing, David Kale, and Greg Ver~Steeg. 2017.
\newblock Anchored correlation explanation: Topic modeling with minimal domain
  knowledge.
\newblock \emph{Transactions of the Association for Computational Linguistics},
  5:529--542.

\bibitem[{Gollapalli and Caragea(2014)}]{cite_text_rank}
Sujatha~Das Gollapalli and Cornelia Caragea. 2014.
\newblock Extracting keyphrases from research papers using citation networks.
\newblock In \emph{Proceedings of the Twenty-Eighth AAAI Conference on
  Artificial Intelligence}, AAAI'14, page 1629–1635. AAAI Press.

\bibitem[{Greene and Cunningham(2006)}]{bbc_news}
Derek Greene and P\'{a}draig Cunningham. 2006.
\newblock Practical solutions to the problem of diagonal dominance in kernel
  document clustering.
\newblock In \emph{Proc. 23rd International Conference on Machine learning
  (ICML'06)}, pages 377--384. ACM Press.

\bibitem[{Jain et~al.(2020)Jain, van Zuylen, Hajishirzi, and Beltagy}]{scirex}
Sarthak Jain, Madeleine van Zuylen, Hannaneh Hajishirzi, and Iz~Beltagy. 2020.
\newblock \href {http://arxiv.org/abs/2005.00512} {Scirex: A challenge dataset
  for document-level information extraction}.
\newblock In \emph{Proceedings of the 58th Annual Meeting of the Association
  for Computational Linguistics}.

\bibitem[{Jin et~al.(2020)Jin, Wei, Du, Xue, and
  Ren}]{hierarchical_attribution}
Xisen Jin, Zhongyu Wei, Junyi Du, Xiangyang Xue, and Xiang Ren. 2020.
\newblock \href {https://openreview.net/forum?id=BkxRRkSKwr} {Towards
  hierarchical importance attribution: Explaining compositional semantics for
  neural sequence models}.
\newblock In \emph{International Conference on Learning Representations}.

\bibitem[{Kennedy et~al.(2020)Kennedy, Jin, Mostafazadeh~Davani, Dehghani, and
  Ren}]{posthocexplain}
Brendan Kennedy, Xisen Jin, Aida Mostafazadeh~Davani, Morteza Dehghani, and
  Xiang Ren. 2020.
\newblock \href {https://doi.org/10.18653/v1/2020.acl-main.483}
  {Contextualizing hate speech classifiers with post-hoc explanation}.
\newblock In \emph{Proceedings of the 58th Annual Meeting of the Association
  for Computational Linguistics}, pages 5435--5442, Online. Association for
  Computational Linguistics.

\bibitem[{Liang et~al.(2021)Liang, Wu, Li, and
  Li}]{liang-etal-2021-unsupervised}
Xinnian Liang, Shuangzhi Wu, Mu~Li, and Zhoujun Li. 2021.
\newblock \href {https://doi.org/10.18653/v1/2021.emnlp-main.14} {Unsupervised
  keyphrase extraction by jointly modeling local and global context}.
\newblock In \emph{Proceedings of the 2021 Conference on Empirical Methods in
  Natural Language Processing}, pages 155--164, Online and Punta Cana,
  Dominican Republic. Association for Computational Linguistics.

\bibitem[{Liu et~al.(2015)Liu, Shang, Wang, Ren, and Han}]{autophrase2}
Jialu Liu, Jingbo Shang, Chi Wang, Xiang Ren, and Jiawei Han. 2015.
\newblock Mining quality phrases from massive text corpora.
\newblock In \emph{Proceedings of the 2015 ACM SIGMOD International Conference
  on Management of Data}, pages 1729--1744.

\bibitem[{Liu et~al.(2019)Liu, Ott, Goyal, Du, Joshi, Chen, Levy, Lewis,
  Zettlemoyer, and Stoyanov}]{roberta}
Yinhan Liu, Myle Ott, Naman Goyal, Jingfei Du, Mandar Joshi, Danqi Chen, Omer
  Levy, Mike Lewis, Luke Zettlemoyer, and Veselin Stoyanov. 2019.
\newblock Roberta: A robustly optimized bert pretraining approach.
\newblock \emph{arXiv preprint arXiv:1907.11692}.

\bibitem[{Liu et~al.(2010)Liu, Huang, Zheng, and Sun}]{Liu2010AutomaticKE}
Zhiyuan Liu, Wenyi Huang, Yabin Zheng, and Maosong Sun. 2010.
\newblock Automatic keyphrase extraction via topic decomposition.
\newblock In \emph{EMNLP}.

\bibitem[{Liu et~al.(2009)Liu, Li, Zheng, and Sun}]{key_cluster}
Zhiyuan Liu, Peng Li, Yabin Zheng, and Maosong Sun. 2009.
\newblock Clustering to find exemplar terms for keyphrase extraction.
\newblock In \emph{Proceedings of the 2009 conference on empirical methods in
  natural language processing}, pages 257--266.

\bibitem[{Luan et~al.(2018)Luan, He, Ostendorf, and Hajishirzi}]{scierc}
Yi~Luan, Luheng He, Mari Ostendorf, and Hannaneh Hajishirzi. 2018.
\newblock \href {https://doi.org/10.18653/v1/D18-1360} {Multi-task
  identification of entities, relations, and coreference for scientific
  knowledge graph construction}.
\newblock In \emph{Proceedings of the 2018 Conference on Empirical Methods in
  Natural Language Processing}, pages 3219--3232, Brussels, Belgium.
  Association for Computational Linguistics.

\bibitem[{Lundberg and Lee(2017)}]{shapley}
Scott~M. Lundberg and Su-In Lee. 2017.
\newblock A unified approach to interpreting model predictions.
\newblock In \emph{Proceedings of the 31st International Conference on Neural
  Information Processing Systems}, NIPS'17, page 4768–4777, Red Hook, NY,
  USA. Curran Associates Inc.

\bibitem[{Mani et~al.(2020)Mani, Yue, Gutierrez, Huang, Lin, and
  Sun}]{clinical_phrase_mining}
Kaushik Mani, Xiang Yue, Bernal~Jimenez Gutierrez, Yungui Huang, Simon Lin, and
  Huan Sun. 2020.
\newblock Clinical phrase mining with language models.
\newblock In \emph{2020 IEEE International Conference on Bioinformatics and
  Biomedicine (BIBM)}, pages 1087--1090. IEEE.

\bibitem[{Marujo et~al.(2013)Marujo, Viveiros, and Neto}]{news}
Luis Marujo, M{\'a}rcio Viveiros, and Jo{\~a}o Paulo da~Silva Neto. 2013.
\newblock Keyphrase cloud generation of broadcast news.
\newblock In \emph{Proceeding of Interspeech 2011: 12th Annual Conference of
  the International Speech Communication Association}.

\bibitem[{Mekala and Shang(2020)}]{application_of_keyphrase1}
Dheeraj Mekala and Jingbo Shang. 2020.
\newblock \href {https://doi.org/10.18653/v1/2020.acl-main.30} {Contextualized
  weak supervision for text classification}.
\newblock In \emph{Proceedings of the 58th Annual Meeting of the Association
  for Computational Linguistics}, pages 323--333, Online. Association for
  Computational Linguistics.

\bibitem[{Mihalcea and Tarau(2004)}]{textrank}
Rada Mihalcea and Paul Tarau. 2004.
\newblock Textrank: Bringing order into text.
\newblock In \emph{Proceedings of the 2004 conference on empirical methods in
  natural language processing}, pages 404--411.

\bibitem[{Montavon et~al.(2017)Montavon, Lapuschkin, Binder, Samek, and
  M{\"u}ller}]{montavon}
Gr{\'e}goire Montavon, Sebastian Lapuschkin, Alexander Binder, Wojciech Samek,
  and Klaus-Robert M{\"u}ller. 2017.
\newblock Explaining nonlinear classification decisions with deep taylor
  decomposition.
\newblock \emph{Pattern Recognition}, 65:211--222.

\bibitem[{Papagiannopoulou and Tsoumakas(2018)}]{reference_vector_algo}
Eirini Papagiannopoulou and Grigorios Tsoumakas. 2018.
\newblock Local word vectors guiding keyphrase extraction.
\newblock \emph{Information Processing \& Management}, 54(6):888--902.

\bibitem[{Papagiannopoulou and Tsoumakas(2019)}]{survey_keyphrase_extraction}
Eirini Papagiannopoulou and Grigorios Tsoumakas. 2019.
\newblock \href {http://arxiv.org/abs/1905.05044} {A review of keyphrase
  extraction}.
\newblock \emph{CoRR}, abs/1905.05044.

\bibitem[{Rajagopal et~al.(2021)Rajagopal, Balachandran, Hovy, and
  Tsvetkov}]{self_explain}
Dheeraj Rajagopal, Vidhisha Balachandran, E.~Hovy, and Yulia Tsvetkov. 2021.
\newblock Selfexplain: A self-explaining architecture for neural text
  classifiers.
\newblock \emph{ArXiv}, abs/2103.12279.

\bibitem[{Ribeiro et~al.(2016)Ribeiro, Singh, and Guestrin}]{Ribeiro2016WhySI}
Marco~Tulio Ribeiro, Sameer Singh, and Carlos Guestrin. 2016.
\newblock "why should i trust you?": Explaining the predictions of any
  classifier.
\newblock \emph{Proceedings of the 22nd ACM SIGKDD International Conference on
  Knowledge Discovery and Data Mining}.

\bibitem[{Rose et~al.(2010)Rose, Engel, Cramer, and Cowley}]{rake}
Stuart Rose, Dave Engel, Nick Cramer, and Wendy Cowley. 2010.
\newblock Automatic keyword extraction from individual documents.
\newblock \emph{Text mining: applications and theory}, 1:1--20.

\bibitem[{Rousseau and Vazirgiannis(2015)}]{partial_match}
Fran{\c{c}}ois Rousseau and Michalis Vazirgiannis. 2015.
\newblock Main core retention on graph-of-words for single-document keyword
  extraction.
\newblock In \emph{European Conference on Information Retrieval}, pages
  382--393. Springer.

\bibitem[{Shang et~al.(2018)Shang, Liu, Jiang, Ren, Voss, and
  Han}]{autophrase1}
Jingbo Shang, Jialu Liu, Meng Jiang, Xiang Ren, Clare~R Voss, and Jiawei Han.
  2018.
\newblock Automated phrase mining from massive text corpora.
\newblock \emph{IEEE Transactions on Knowledge and Data Engineering},
  30(10):1825--1837.

\bibitem[{Shrikumar et~al.(2017)Shrikumar, Greenside, and Kundaje}]{shrikumar}
Avanti Shrikumar, Peyton Greenside, and Anshul Kundaje. 2017.
\newblock Learning important features through propagating activation
  differences.
\newblock In \emph{International Conference on Machine Learning}, pages
  3145--3153. PMLR.

\bibitem[{Sterckx et~al.(2015)Sterckx, Demeester, Deleu, and
  Develder}]{Sterckx2015WhenTM}
Lucas Sterckx, Thomas Demeester, Johannes Deleu, and Chris Develder. 2015.
\newblock When topic models disagree: Keyphrase extraction with multiple topic
  models.
\newblock \emph{Proceedings of the 24th International Conference on World Wide
  Web}.

\bibitem[{Sun et~al.(2020)Sun, Qiu, Zheng, Wang, and Zhang}]{Sun2020SIFRankAN}
Yi~Sun, Hangping Qiu, Yu~Zheng, Zhongwei Wang, and Chaoran Zhang. 2020.
\newblock Sifrank: A new baseline for unsupervised keyphrase extraction based
  on pre-trained language model.
\newblock \emph{IEEE Access}, 8:10896--10906.

\bibitem[{Tomokiyo and Hurst(2003)}]{ngram_lm_keyphrase}
Takashi Tomokiyo and Matthew Hurst. 2003.
\newblock \href {https://doi.org/10.3115/1119282.1119287} {A language model
  approach to keyphrase extraction}.
\newblock In \emph{Proceedings of the ACL 2003 Workshop on Multiword
  Expressions: Analysis, Acquisition and Treatment - Volume 18}, MWE '03, page
  33–40, USA. Association for Computational Linguistics.

\bibitem[{Wallach(2006)}]{Wallach2006TopicMB}
Hanna~M. Wallach. 2006.
\newblock Topic modeling: beyond bag-of-words.
\newblock \emph{Proceedings of the 23rd international conference on Machine
  learning}.

\bibitem[{Wan and Xiao(2008)}]{single_rank}
Xiaojun Wan and Jianguo Xiao. 2008.
\newblock Single document keyphrase extraction using neighborhood knowledge.
\newblock In \emph{AAAI}, volume~8, pages 855--860.

\bibitem[{Wang et~al.(2019)Wang, Gao, Huang, and
  Zhou}]{application_of_keyphrase4}
Wenbo Wang, Yang Gao, He-Yan Huang, and Yuxiang Zhou. 2019.
\newblock Concept pointer network for abstractive summarization.
\newblock In \emph{Proceedings of the 2019 Conference on Empirical Methods in
  Natural Language Processing and the 9th International Joint Conference on
  Natural Language Processing (EMNLP-IJCNLP)}, pages 3076--3085.

\bibitem[{Yang et~al.(2019)Yang, Dai, Yang, Carbonell, Salakhutdinov, and
  Le}]{xlnet}
Zhilin Yang, Zihang Dai, Yiming Yang, Jaime Carbonell, Russ~R Salakhutdinov,
  and Quoc~V Le. 2019.
\newblock \href
  {https://proceedings.neurips.cc/paper/2019/file/dc6a7e655d7e5840e66733e9ee67cc69-Paper.pdf}
  {Xlnet: Generalized autoregressive pretraining for language understanding}.
\newblock In \emph{Advances in Neural Information Processing Systems},
  volume~32. Curran Associates, Inc.

\bibitem[{Yu and Ng(2018)}]{wikirank}
Yang Yu and Vincent Ng. 2018.
\newblock Wikirank: Improving keyphrase extraction based on background
  knowledge.
\newblock \emph{arXiv preprint arXiv:1803.09000}.

\bibitem[{Zhang et~al.(2022)Zhang, Chen, Wang, Deng, Zhang, Li, Wang, and
  Cao}]{zhang-etal-2022-mderank}
Linhan Zhang, Qian Chen, Wen Wang, Chong Deng, ShiLiang Zhang, Bing Li, Wei
  Wang, and Xin Cao. 2022.
\newblock \href {https://doi.org/10.18653/v1/2022.findings-acl.34} {{MDER}ank:
  A masked document embedding rank approach for unsupervised keyphrase
  extraction}.
\newblock In \emph{Findings of the Association for Computational Linguistics:
  ACL 2022}, pages 396--409, Dublin, Ireland. Association for Computational
  Linguistics.

\end{thebibliography}
\bibliographystyle{acl_natbib}

\appendix

\section{Appendix}
\label{sec:appendix}


\begin{table*}[t]
    \centering
    \small
    \begin{tabular}{c|l}
        S.No. & Top words from removed topic \\
        \hline
        1 & proposed;propose novel;propose;proposed method;method\\
        2 & generalization;study;analysis;suggest;provide\\
        3 & outperforms;existing;existing methods;outperforms stateoftheart;methods\\
        4 & state;art;state art;shortterm;current state\\
        5 & effectiveness;demonstrate effectiveness;source;effectiveness proposed;student\\
        6 & training;training data;training set;training process;model training\\
        7 & experimental;experimental results;results;results demonstrate;experimental results demonstrate\\
        8 & experiments;extensive;extensive experiments;experiments demonstrate;conduct\\
        9 & performance;improves;significantly;improve;improved\\
        10 & recent;shown;recent work;recent advances;success\\
        11 & achieves;introduce;competitive;achieves stateoftheart;introduce new\\
        12 & trained;model trained;models trained;networks trained;trained using\\
        13 & present;paper present;present novel;work present;monte\\
        14 & widely;parameters;widely used;proposes;paper proposes\\
        15 & simple;benchmark datasets;benchmark;propose simple;simple effective\\
        16 & prior;approach;sampling;continuous;prior work\\
        17 & program;introduces;programs;future;paper introduces\\
        18 & solve;challenging;able;complex;challenging problem\\
        19 & challenge;current;challenges;open;current stateoftheart\\
        20 & rate;good;good performance;l;regime\\
        21 & works;previous works;existing works;focus;scenarios\\
        22 & evaluate;evaluation;tackle;tackle problem;evaluate method
    \end{tabular}
    \caption{22 Generic topics removed from the 75 topic labels learned using topic modeling on ICLR data.}
    \label{tab:removed_topics}
\end{table*}
\subsection{Evaluation Datasets}
\label{app:datasets}

\textbf{SemEval-2017} \cite{semeval2017} consists of 500 abstracts taken from 12 AI conferences covering Computer Science, Material Science, and Physics. The entities are annotated with Process, Task, and Material labels, which form the fundamental concepts in scientific literature. Identification of the keyphrases was  subtask A of the ScienceIE SemEval task \cite{augenstein-etal-2017-semeval}.

\textbf{SciERC} \cite{scierc} extends SemEval-2017 by annotating more entity types, relations, and co-reference clusters to include broader coverage of general AI. 
The dataset was annotated by a single domain expert who had high (76.9\%) agreement with three other expert annotators on 12\% subset of the dataset. 

\textbf{SciREX} \cite{scirex} is a document-level information extraction dataset, covering entity identification and n-ary relation formation using salient entities. Human and automatic annotations were used to annotate 438 full papers with salient entities, with a distant supervision from the Papers With Code\footnote{\url{https://paperswithcode.com/}} corpus. This dataset can help verify the performance of models on full papers.

\textbf{500N-KPCrowd} \cite{news} is a keyphrase extraction dataset in the news domain. This data consists of 500 articles from 10 topics annotated by multiple Amazon Mechanical Turk workers for important keywords. Following the baselines on this datasets, we pick keywords that were among the top two most frequently chosen by the human annotators. Since no span-level information for these keywords is given, we annotate all occurrences of the chosen keywords in the document to obtain a list of span labels, which we use to evaluate all the models.   

\subsection{Implementation Details}
\label{app:impl}
Here, we present the hyper-parameters for all experiments along with their corresponding search space. We chose all hyperparameters based on the development set performance on the SciERC dataset. We considered RoBERTa \cite{roberta} and XL-NET \cite{xlnet} based encoders and finally chose RoBERTa for faster compute times. We experimented with learning-rates from the set of 1e-5,2e-5,5e-5,1e-4 and 2e-4. We chose 2e-5 as the final learning rate. Our batch size of 8 was chosen after experimenting with 4, 8, 12 and 16. 
The size of the weights matrix in the classification layer was chosen to be 64 from a set of 16,32,64 and 128. 
The $\alpha$ parameter used for regularization was fixed at 0.5. We tried values between 0.1 and 0.9 and did not find signifcant difference. We saved the model based on best weighted F1 on the topic prediction task. All training runs took less than 3 hours on 2 Nvidia 2080Ti GPUs, except on the ICLR dataset, which took 8 hours. All results are from a single run.

\end{document}